\documentclass{article}
\usepackage{spconf,amsmath,graphicx}
\usepackage{xcolor}
\usepackage{hyperref}
\usepackage{algorithm}
\usepackage{algpseudocode}

\title{Joint ANN-SNN Co-training for Object Localization and Image
  Segmentation}

\name{Marc Baltes$^1$, Nidal Abujahar$^1$, Ye Yue$^1$, Charles
  D. Smith$^2$, Jundong Liu$^1$\thanks{Corresponding author:
    Dr. Jundong Liu. Email: liuj1@ohio.edu.
}}  \address{ $^1$School
  of Electrical Engineering and Computer Science \\Ohio University,
  Athens OH 45701 \\ $^2$Department of Neurology \\ University of
  Kentucky, Lexington KY 40506}
\begin{document}
%
\maketitle
\begin{abstract}
The field of machine learning has been greatly transformed with the
advancement of deep artificial neural networks (ANNs) and the
increased availability of annotated data. Spiking neural networks
(SNNs) have recently emerged as a low-power alternative to ANNs due to
their sparsity nature.

In this work, we propose a novel hybrid ANN-SNN co-training framework
to improve the performance of converted SNNs. Our approach is a
fine-tuning scheme, conducted through an alternating, forward-backward
training procedure. We apply our framework to object detection and
image segmentation tasks. Experiments demonstrate the effectiveness of
our approach in achieving the design goals.

\end{abstract}

\section{Introduction}


Over the past decade, deep artificial neural networks (ANNs) have
revolutionized many AI-related fields, including computer vision,
natural language processing, human speech processing, and
autonomy. This remarkable progress
is in part due to the emerging of large amount of annotated data and
the widespread availability of high-performance computing devices and
general-purpose graphics processing units (GPUs).
However, with this comes a huge computational and power burden, which
limits the application of ANNs in many tasks that require low size,
weight and power (SWaP) devices.  Spiking neural networks (SNNs) have
recently emerged as a low-power alternative,
whose neurons imitate the temporal and sparse spiking nature of
biological neurons \cite{roy2019towards,davies2021advancing}.
SNN neurons consume energy only when spikes are generated,
leading to sparser activations and natural gains in SWaP.

%
%
 An SNN can be obtained by either converting from a fully trained ANN,
 or through a direct training procedure where a surrogate gradient
 needs to be used for the network to conduct backpropagation
 \cite{lee2016training}. ANN-to-SNN conversion methods, while
 effective, often require sophisticated thresholding or normalization
 procedures to determine the spiking neuron configurations
 \cite{smith2021evaluation,diehl2015_conversion,rueckauer2016theory,
   snn-max2019,
   yue2023snn}. Direct training solutions commonly suffer from
 expensive computation burdens on complex network architectures
 \cite{shrestha2018slayer, wu2018spatio,
   rathi_fine_tuning2020,li2021differentiable}.
%
 
Recently, hybrid training methods have been proposed, aiming to
combine the power of ANN-SNN conversion and direct SNN training.
Wu et al. \cite{wu_tandem2021} proposed a TANDEM framework for
training of an ANN and SNN jointly.  Rathi et
al. \cite{rathi_fine_tuning2020} developed a hybrid SNN training
approach, in which backpropagation is used to fine-tune the network
after conversion. Significant performance improvements have been
reported. This solution, however, requires rather sophisticated
normalization setups. 
 

In addition, most of the current SNN models focus on  
image recognition related tasks, mostly through convolutional neural
network (CNN) models.  Other important tasks, such as object detection
\cite{spike_yolo2020} and image segmentation
\cite{kim2021beyond,patel2021spiking} have not been widely studied. It
should be noted that the former is a regression task, which requires a
network to work on real numbers. Image segmentation, on the other
hand, requires the network to produce dense classifications at the
pixel level. Both requirements pose challenges for spiking networks.


In this work, we propose a new hybrid ANN-SNN training scheme to
address the aforementioned issues. Our approach is a fine-tuning
scheme: an ANN will be trained first, before being converted to an
SNN; then the weights of the SNN will be updated through an
alternating, forward-backward training procedure. The forward
propagation carries signals in spike-train format, essentially
conducting an SNN inference. The backward passes use ANN
backpropagation to update the network weights.

We build our networks based on soft {\it leaky integrat-and-fire} 
neurons,
which makes ANN-to-SNN switching rather straightforward.
%
%
We apply the proposed hybrid ANN-SNN fine-tuning scheme to object
detection and image segmentation tasks. To the best of our knowledge,
this is the first hybrid SNN training work proposed on these two
tasks.


\section{Method}

In this section, we will introduce the proposed hybrid ANN-SNN
training scheme, as well as its applications in object localization
and image segmentation.  We start with the ANN models for the two
tasks.

 \subsection{Baseline ANN Models}




{\bf Object localization} is a common step in object detection
algorithms to obtain accurate bounding boxes for the objects of
interest. In the R-CNN model \cite{girshick2014rich}, multiple region
proposals are first extracted from an input image and then sent to a
pre-trained CNN to extract features.
A support vector machine (SVM) model takes the extracted features to
classify the objects inside, followed by a bounding box regression step
to localize the objects. 

\begin{figure}[htb]
  \centering \centerline{\includegraphics[width=8.5cm]{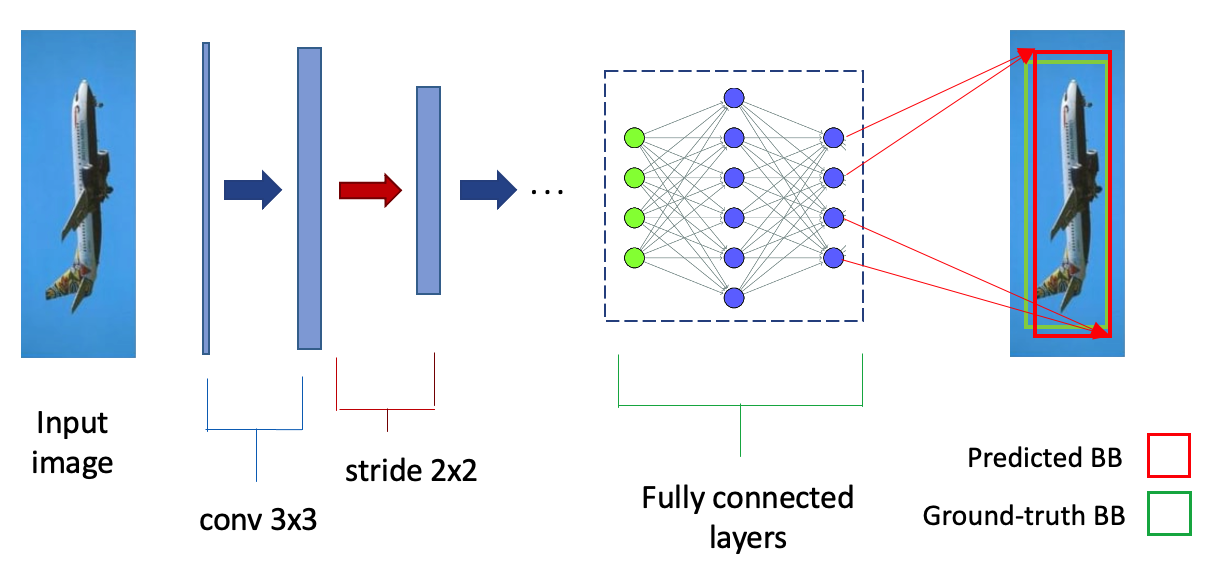}}
\caption{An illustration of the network structure of {\it LocNet}. {\it BB}
  stands for bounding box. Refer to text for details. }
\label{fig:cae}
\end{figure}

In this work, we focus on the localization step of the R-CNN model,
with the goal to identify an accurate bounding box to capture the
object in the input image. Our solution is built as a CNN, as shown in
Fig.~\ref{fig:cae}.  It starts with a convolutional layer, followed by
a pooling layer. The pooling is implemented using a convolution
operation with stride 2. We choose not to use max-pooling as it is
difficult to implement in spiking networks. This convolution-pooling
pair is repeated three times, followed by four fully connected
layers. The network produces the predictions for the corner
coordinates (totally 4 real-valued numbers) of the bounding box.
We name this baseline model {\it LocNet}. 

{\bf Segmentation network}
%
%
We use a convolutional autoencoder (CAE) as the baseline segmentation
network, which follows an encoding and decoding architecture.
Taking 2D images as inputs, the encoding part
repeats the conventional {\it convolution + pooling} layers to extract
high-level latent features. The decoding part reconstructs the
segmentation ground truth mask by using {\it transpose / deconvolution}
layers.

Our CAE model can be regarded as a simplified version of U-Net
\cite{ronneberger2015u}, which is a popular
solution for medical image segmentation. Modifications have been made
to suit our data and task. First, we reduce the number of
convolutional layers to 10 to fit the image size of our data.
Second, we replace max pooling layers with average-pooling, as there
is no effective implementation of max pooling in SNNs.  Moreover, we
remove the concatenation operations (skip connections) that connect the
encoding and decoding stacks.
This modification is
aimed to avoid complicated data synchronization between encoder and
decoder layers in the converted SNN model.


\subsection{Proposed hybrid SNN-ANN co-training}

Our proposed ANN and SNN networks are developed under
NengoDL \cite{rasmussen2019nengodl}, which provides a number of
spiking and non-spiking neurons.
NengoDL also provides a variety of controls for SNNs,
including neuron activations, synaptic smoothing applied to each
connection, firing rate of each layer, and time steps that images are
presented in the model.



\begin{figure}[htb]
  \centering \centerline{\includegraphics[width=8.5cm]{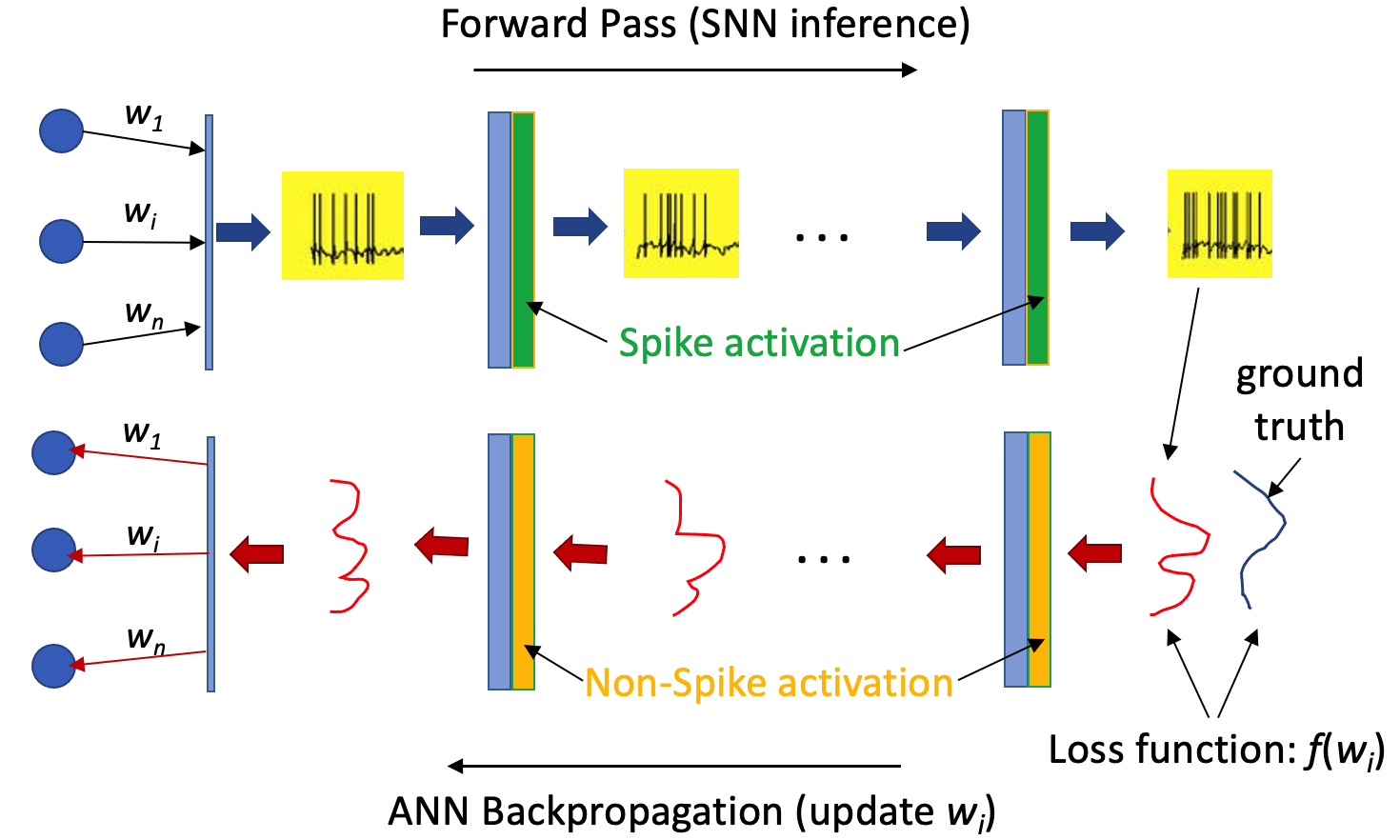}}
  \caption{An illustration of the proposed hybrid ANN-SNN
    co-training solution. Refer to text for details. }
\label{fig:hybrid}
\end{figure}

Our hybrid ANN-SNN co-training approach is illustrated in
Fig.~\ref{fig:hybrid}.  An ANN of the task of interest will be trained
first, before being converted to an SNN. Then, the weights of the
converted SNNs are updated through an alternating forward-backward
fine-tuning procedure. The forward pass is essentially an SNN
inference procedure, which uses quantized activation functions
(green-color neurons in Fig.~\ref{fig:hybrid}) to generate spike
trains, flowing through the network. The backward pass computes the
error between SNN output and the ground-truth, and uses the gradient
of the loss w.r.t. the weights to update the network parameters. Spike
activation functions are switched to corresponding non-spike functions
along the backward pass.



In our LocNet model, we use {\it soft leaky integrate-and-fire} (LIF)
neurons proposed by Hunsberger and Eliasmith
\cite{nengodl_softlif2015, rasmussen2019nengodl}.  Extended from LIF,
soft LIF neurons have smoothing operations applied around the firing
threshold. As explained in \cite{nengodl_softlif2015}, the LIF neuron
dynamics are defined by the following equation:

\begin{equation}
\tau_{RC} \frac{dv(t)}{dt} = -v(t) + J(t)
\end{equation}

where the membrane voltage is $v(t)$, the input current is $J(t)$, and
$\tau_{RC}$ represents the membrane time constant. The neuron fires a
spike when the voltage $V_{th} = 1$. The voltage is then set and
remains at zero until a refractory period of $\tau_{ref}$ has
passed. To solve for the steady-state firing rate the authors set a
constant input current given by $J(t) = j$ which produces this
equation:

\begin{equation}
r(j) = 
 \begin{bmatrix}
\tau_{ref} - \tau_{RC} \log(1 + \frac{V_{th}}{\rho(j - V_{th})})
\end{bmatrix}^{-1}
\end{equation}

where $\rho(x) = \max(x, 0)$. However, as $j \rightarrow 0_+$, the
steady-state equation's derivative approaches infinity. To counteract
this, the authors set $\rho(x) = \gamma \log(\begin{bmatrix} 1
+ e^{x/\gamma}
\end{bmatrix})$. This allows for control over the smoothing that is applied where
$\rho(x) \rightarrow \max(x, 0)$ as $\gamma \rightarrow 0$. As a
result from the smoothing, gradient-based backpropagation can now be
carried out to train the network. This design makes the conversion
from ANN to SNN rather straightforward, removing the need of
complicated thresholding or weight normalization steps in many
conversion algorithms. In this work, we take advantage of the
convenience brought by this soft LIF model to switch between ANN and
SNN to fine-tune both. For our LocNet, we use the mean squared error
(MSE) function as the network loss.



In our CAE model, we use the {\it rectified linear} neurons in
NengoDL, whose activity scales linearly with the current, unless
it passes below zero, at which point the neural activity will stay at
zero.  It should be noted SNNs' currents are either positive or zero.
The gradient of rectified linear neurons is 1 for positive currents
and 0 for when the current equals zero. The loss function for our CAE
model is set to a weighted combination of binary cross entropy (BCE)
and the Dice loss. The contribution is set to 0.5 for each loss
component.


NengoDL’s SNN simulator was used to run models as SNNs and collect
execution results. When using the simulator, we switch the non-spiking
neurons to their spiking counterparts. In LocNet, the soft LIF neurons
are switched to spiking LIF neurons while in the CAE model the
Rectified Linear neurons are replaced by the Spiking Rectified Linear
version. In order to make fair and consistent comparisons,
the hyperparameters were maintained the same for the ANN, SNN and hybrid
ANN-SNN of the same network.




\section{Implementation details}



{\bf The data} for object localization is the airplane subset of Caltech-101
\cite{caltech_data}.
There are in total 800 images of airplanes with the size of
300$\times$200 pixels.
Each image is accompanied with four real-valued numbers that make up
the bounding box of the airplane.


The image segmentation task in this work is to extract human brain
Hippocampi from magnetic resonance images (MRIs).  The data used were
obtained from the ADNI database (\url{http://adni.loni.usc.edu}), and
extracted in our previous work
\cite{chen2017hippocampus,chen2017accurate}. In
total, 110 base-line T1-weighted whole brain MRI images from different
subjects along with their hippocampus masks were downloaded.
Due to the fact
that the hippocampus only occupies a very small part of the whole
brain and its position in the whole brain is relatively fixed, we
roughly cropped the right face of the hippocampus of each subject and use them as
the input for segmentation.  The size of the cropping box is $24
\times 56 \times 48$.

\subsection{Training and testing}

%
%
In SNNs, {\it firing rates} play an important role in determining the
total energy consumption in the model as well as how information is
passed down throughout the network.
%
%
%
NengoDL provides two different ways to set the firing rates at network
layers: {\it post-training scaling} and {\it regularizing during
  training}.  The former allows us to apply a linear scale to the
inputs of all the neurons and then scale down the outputs with the
same rate. The latter is used to directly optimize the firing rates towards
certain target values we specify.
{\bf LocNet training} In the experiments of LocNet,
we randomly chose 90\% of the data for
training while the other 10\% were used for testing. Our ANN model
was trained with NengoDL's soft LIF neurons with a smoothing factor of 0.005.
The firing rates were {\it regularized during training} with a target
rate of 250 Hz. Different weights were assigned to the layers in the
regularization. As the output layer is the most important, we set its
weight to 1. Other layers are less important so we set each of their
loss weights to 0.01. Our ANN model was trained for 50 epochs using
the Adam optimizer with a learning rate of 0.01.

For hybrid training, the same firing rate regularization technique was
used to keep the neurons firing at around 250 Hz.
The soft LIF neurons in ANN were replaced with our hybrid soft LIF
neurons. In addition to this change, a synapse of 0.005 seconds was
added to all connections in the network.
The hybrid model was trained for 7 epochs which was determined by
an early stopping process.
The learning rate was also decreased to 0.001.


{\bf CAE training} Our dataset to train our CAE network consists of
5280 2D images and masks of the right face of the hippocampus. 4780 of
these were used for training while 500 were used for testing. The ReLU
activations were swapped out for NengoDL's Rectified Linear neurons
during training. Firing rate scaling was used on these neurons with a
scaling factor of 1000. The ANN was trained for 100 epochs using the
Adam optimizer and the learning rate was set to 0.00001.



In the hybrid training step, the Rectified Linear neurons were
replaced with our hybrid Rectified Linear neurons.
A {\it post-training scaling} factor of 1000 was also used on both
neuron types during training to keep results consistent.  A synaptic
filter of 0.005 seconds was added to the output of all spiking
neurons. Finally, an early stopping technique was used and trained the
model for an additional 13 epochs.
%


\section{Experiments and Results}

In this section, we present and evaluate the experimental results for
the proposed models. 
%
\subsection{Object localization}
 {\it Intersection over Union} (IoU) is used to evaluate the object
 localization networks.
IoU measures how closely two sets of elements overlap, and it is
defined as the ratio of the overlap area to the combined area of
prediction and ground-truth.
%

Three network models are compared in this experiment. The first is the
{\it pure ANN} (LocNet) model, which is built and trained with
standard CNN components. The second is the SNN directly converted from
the ANN ({\it converted SNN}).  The third one is our {\it hybrid SNN},
which is fine-tuned upon the converted SNN with our hybrid training
approach. The three models use the same hyperparameters. For SNN
models, the soft LIF rate neurons are switched to spiking LIF neurons
during testing. {\it Firing rate regularization} is used to identify
optimal firing rate for the two SNN models.


\begin{table}[h]
\centering
\begin{tabular}{ |c|c|c| } 
  \hline Model & Mean IoU & Avg. Firing Rate \\ \hline \hline {\it
  pure ANN} (LocNet) & 0.8945 & N/A \\ \hline {\it converted SNN} & 0.7431 &
  174.5583 Hz \\ \hline {\it hybrid SNN} & \textbf{0.8253}
  & \textbf{137.5238} Hz \\ \hline
\end{tabular}
\caption{Results from the three localization models.}
\label{rcnn_table}
\end{table}


Table \ref{rcnn_table} shows the results obtained from the three
models. The baseline ANN model achieves a mean IoU of 0.8945. When
converted to SNN, the network accuracy drops to 0.7278. Our hybrid
fine-tuning procedure is able to improve the performance, measured by
IoU, to 0.8253. In addition, our hybrid SNN obtains a slightly smaller
average firing rate than the converted version, which poses an
advantage from the energy consumption perspective.  We believe this is
partly due to the weighted regularization terms that are added to the
loss function for each layer in the network. The regularization urges
the network to implicitly learn the scaling factor to the weights
instead of applying a post-scaling in the inference stage for the
SNN. Learning an optimized scaling factor for each layer results in
optimizing the firing rates.

\subsection{Hippocampus segmentation}


Dice ratio is used as the performance metric to measure the overlap
between the automatic segmentation and the ground truth mask. The
results of the competing models are shown in
Table~\ref{cae_table}. Similar to the previous experiment, we have
{\it pure ANN}, {\it converted SNN}, and {\it hybrid SNN} as the
competing models.  All networks use the same set of
hyper-parameters. For the firing rates, we use the {\it post-training
scaling} approach and empirically set the same scaling factor of 1000
for all neurons. We chose not to use {\it firing rate regularization}
due to the difficulty of setting up appropriate contribution weights.

The baseline {\it pure ANN} model was able to achieve a Dice
coefficient of 0.8221. The converted model, however, suffers from a
large performance loss, only achieving a Dice score of 0.4859. After
going through the hybrid fine-tuning process, the hybrid SNN model was
able to bring the Dice score back to 0.7648.
This clearly demonstrates the effectiveness of the proposed hybrid
fine-tuning strategy.

\begin{table}[h]
\centering
\begin{tabular}{ |c|c|c| } 
  \hline Model & Dice & Avg. Firing Rate \\ \hline \hline {\it
  pure ANN} (\textrm{CAE}) & 0.8221 & N/A \\ \hline {\it converted SNN} & 0.4859 &
  708.2347 Hz \\ \hline {\it hybrid SNN} & \textbf{0.7648} & 731.3800
  Hz \\ \hline
\end{tabular}
\caption{Results from the three segmentation models. }
\label{cae_table}
\end{table}

\begin{figure}
  
\begin{minipage}[b]{0.48\linewidth}
  \centering
  \centerline{\includegraphics[width=3.5cm]{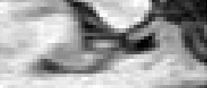}}
  \centerline{(a) Input slice}\medskip
\end{minipage}
\begin{minipage}[b]{0.48\linewidth}
\centering
  \centerline{\includegraphics[width=3.5cm]{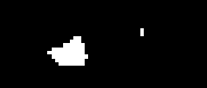}}
  \centerline{(b)  Ground-truth}\medskip
\end{minipage}
\begin{minipage}[b]{.32\linewidth}
  \centering
  \centerline{\includegraphics[width=2.7cm]{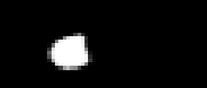}}
  \centerline{(c)  ANN}\medskip
\end{minipage}
\hfill
\begin{minipage}[b]{0.32\linewidth}
  \centering
  \centerline{\includegraphics[width=2.7cm]{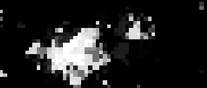}}
  \centerline{(d) Converted SNN}\medskip
\end{minipage}
\hfill
\begin{minipage}[b]{0.32\linewidth}
  \centering
  \centerline{\includegraphics[width=2.7cm]{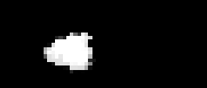}}
  \centerline{(e) Hybrid SNN}\medskip
\end{minipage}
\caption{An example of input slice and segmentation results. }
\label{fig:masks}
\end{figure}

To visualize the effects of hybrid fine-tuning, we look into a number
of slices and compare the masks generated from the three
models. Fig.~\ref{fig:masks} shows a typical example, where (a) and
(b) are the input slice and the ground-truth
mask. Fig~\ref{fig:masks}.(c), (d) and (e) show the network outputs
(probabilistic masks) for ANN, converted SNN and hybrid SNN,
respectively. The mask from the ANN matches the ground-truth rather
well. The mask from converted SNN, however, contains many artifacts,
leading to a poor Dice score after thresholding. The hybrid
fine-tuning is able to correct the artifacts and significantly improve
the segmentation performance for the SNN model.




%
%

\section{Conclusions}

In this work, we present a hybrid ANN-SNN fine-tuning scheme. Our
approach is fairly general and can potentially be applied to many
convolutional networks implemented using the soft LIF neurons or the
rectified linear neurons. We take object localization and image
segmentation as testbed applications, and the effectiveness of our
approach is well demonstrated.  Exploring more network applications,
as well as developing new co-training solutions are our next steps.

\bibliographystyle{IEEEbib}
\bibliography{seg,snn}

\end{document}